# POSITION ANALYSIS OF THE RRP-3(SS) MULTI-LOOP SPATIAL STRUCTURE (*)

**Carlo Innocenti**

Department of Mechanical and Civil Engineering
University of Modena and Reggio Emilia
via Vignolese, 905 – 41100 Modena
Italy
carlo.innocenti@unimore.it

**Philippe Wenger**

Institut de Recherche en Communications et
Cybernetique de Nantes
1, rue la Noe – BP 92101 44321 Nantes Cedex 3
France
philippe.wenger@irccyn.ec-nantes.fr

**ABSTRACT**

The paper presents the position analysis of a spatial structure composed of two platforms mutually connected by one RRP and three SS serial kinematic chains, where R, P, and S stand for revolute, prismatic, and spherical kinematic pair respectively. A set of three compatibility equations is laid down that, following algebraic elimination, results in a $28^{th}$-order univariate algebraic equation, which in turn provides the addressed problem with 28 solutions in the complex domain.

Among the applications of the results presented in this paper is the solution to the forward kinematics of the Tricept, a well-known in-parallel-actuated spatial manipulator.

Numerical examples show adoption of the proposed method in dealing with two case studies.

## INTRODUCTION

Parallel kinematics machines have gained more and more interest in research as well as in industry. In effect, their high stiffness and low inertia make them attractive alternative designs for demanding tasks such as high speed machining [1]. Parallel kinematics machines may feature a fully-parallel kinematic architecture, like the Gough-Stewart platform, or a hybrid parallel-serial kinematic architecture, like the Tricept machine, in which a parallel positioning device carries a serial wrist [2].

With reference to Fig.1, a Tricept manipulator can be thought of as obtained by serially connecting two elemental manipulators, precisely a three-degree-of-freedom (3-dof) in-parallel-actuated manipulator and – depending on the specific Tricept model – a 2-dof or 3-dof serial manipulator (Fig. 1 shows a 6-dof Tricept).

The in-parallel-actuated elemental manipulator of a Tricept consists of a fixed base connected to a movable platform by four serial kinematic sub-chains. One of these sub-chains is of type UP (U stands for universal – or Hooke – joint, P means prismatic kinematic pair), whereas the remaining three sub-chains are of type UPS (S signifies spherical kinematic pair). The serial elemental manipulator connects the already-mentioned movable platform to the end-effector of the Tricept via a universal joint (5-dof Tricept) or a spherical wrist realized by three revolute pairs having mutually-intersecting axes (6-dof Tricept, see Fig. 1).

In any Tricept, the centers of the U-joints in the UPS sub-chains, as well as the centers of the S-joints, are at the vertices of equilateral triangles fixed to the base and to the movable platform respectively. Moreover, the center of the U-joint in the UP sub-chain is equidistant from the centers of the remaining U-joints; in the U-joint of the UP sub-chain, the axis of the revolute kinematic pair next to the manipulator base is either orthogonal or parallel to a side of the aforementioned fixed equilateral triangle; the line through the center of the U-joint of the UP sub-chain and parallel to the sliding movement of this sub-chain's P-joint intersects the movable triangle at its center; the UP sub-chain is so arranged as to make – for a subset of manipulator configurations – the sides of the movable equilateral triangles parallel to and equidistant from the sides of the fixed equilateral triangle.

The Tricept end-effector is moved in space by actuating the P pairs in the three UPS sub-chains of the in-parallel-actuated elemental manipulator, as well as the revolute pairs of the serial elemental manipulator (in Fig. 1 all actuated kinematic pairs are highlighted by asterisks). In the sequel, only the case of a 6-dof Tricept will be taken into account, the differences with respect to the case of a 5-dof Tricept being manifest and easily manageable.

Position analysis is a necessary step for the control of a manipulator. The position analysis problem consists in determining the relationship between the parameters of motion of the actuated joints and the rigid-body position (location) of the manipulator's end-effector. The position analysis problem comprises two dual problems, namely, the inverse kinematics and the forward kinematics.

The inverse kinematics of a manipulator is the search for the parameters of motions of all actuated kinematic pairs in the UPS sub-chains once the location of the end-effector has been assigned. It is not difficult to recognize that the inverse kinematics of the in-parallel-actuated elemental manipulator of a Tricept is equivalent to the inverse kinematics of a serial UP regional manipulator endowed with a spherical wrist [3]. Addressing exhaustively this problem leads to eight real sets of parameters of motion that stem from solving first- and second-order univariate algebraic equations only.

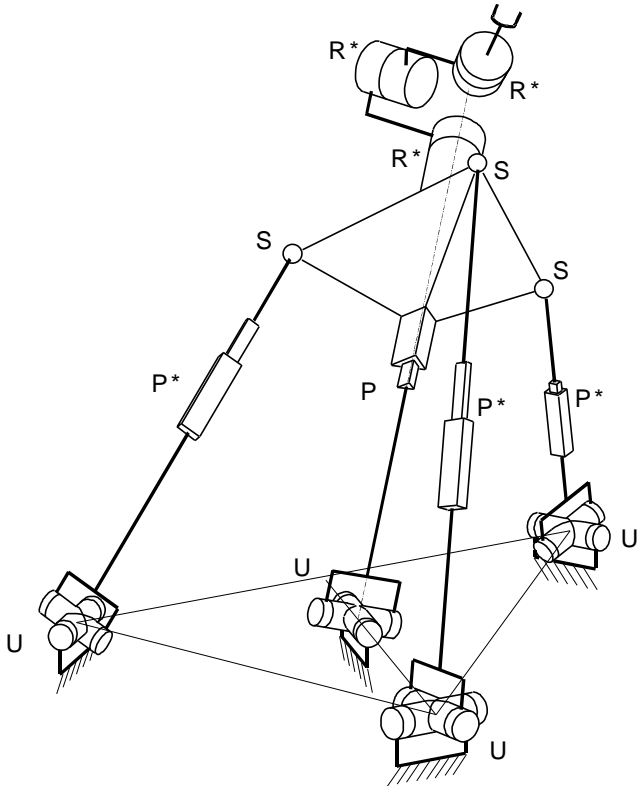

**Figure 1** – A six-degree-of-freedom Tricept manipulator.

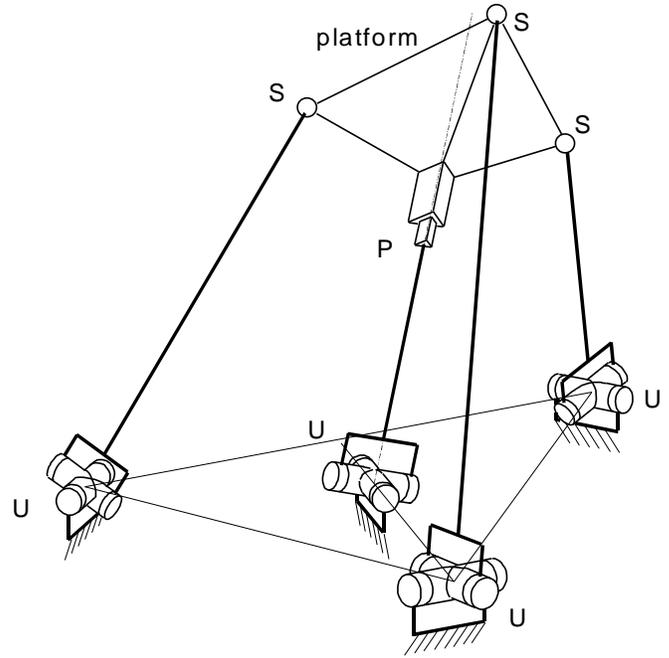

**Figure 2** – The UP-3(US) structure associated with a Tricept manipulator (type-I).

Conversely, the forward kinematics of a manipulator consists in finding the possible locations of the end-effector once the variables of motions of all actuated kinematic pairs have been prescribed. Solving the forward kinematics of a parallel manipulator is a difficult task: despite the copious literature on this topic, no general tool exists that is applicable to any parallel manipulator [4]. For a Tricept, a convenient way of tackling the problem consists in solving separately the forward kinematics of the two aforementioned elemental manipulators. The forward kinematics of the in-parallel-actuated elemental manipulator is equivalent to searching for all possible locations of the platform that are compatible with the assigned parameters of motion of the three actuated prismatic pairs in the UPS sub-chains. For the serial elemental manipulator, on the other hand, the forward kinematics consists in finding the only location of the Tricept end-effector – relative to the movable platform – that results from a given choice of the parameters of motion of the three revolute pairs. As soon as the outcomes of both forward kinematics problems become available, answering the forward kinematics of the Tricept turns out to be a trivial task.

Solving the forward kinematics of the in-parallel-actuated elemental manipulator of a Tricept is equivalent to finding the assembly configurations of the structure that results from the elemental manipulator itself by freezing the three actuated prismatic pairs. Such a UP-3(US) structure – shown in Fig. 2 – has three links each joined to base and platform by a universal joint and a spherical pair respectively. It can be easily recognized that the location of the platform does not change if at the extremities of these links the universal joint and the spherical pair swap places, or even if the universal joint is replaced by a spherical pair. Therefore the forward kinematics of the in-parallel-actuated elemental manipulator of a Tricept can even be solved by looking for the assembly configurations of the linkage represented in Fig. 3.

Despite the freedom of any SS connecting rod to revolve about the line through the extremity spherical pair centers, the UP-3(SS) linkage in Fig. 3 will be conventionally termed a structure because what is relevant here is that the connectivity [5] of the platform relative to the base is still zero as in the original UP-3(US) structure. Should the axes of the revolute pairs in the universal joint of the UP-3(SS) structure become mutually skewed and inclined at a generic angle, the direction of sliding motion in the prismatic kinematic pair be free to form a generic angle with the direction of the axis of the adjacent revolute pair, and the centers of all spherical joints be generically placed on the base and the platform, then the UP-3(SS) structure of Fig. 3 would turn into the RRP-3(SS) general-geometry structure represented in Fig. 4 (character R in structure's acronym means revolute kinematic pair).

The scope of this paper is to present a method for solving in polynomial form the position analysis of the general-geometry RRP-3(SS) spatial structure (see Fig. 4). Starting from a set of three compatibility equations in three unknowns, the method presented in this paper reduces the solution of the position analysis to finding the roots of a $28^{th}$-order univariate algebraic equation. Each root of this equation corresponds to an assembly configuration of the RRP-3(SS) structure, which means that 28 assembly configurations exist in the complex domain.

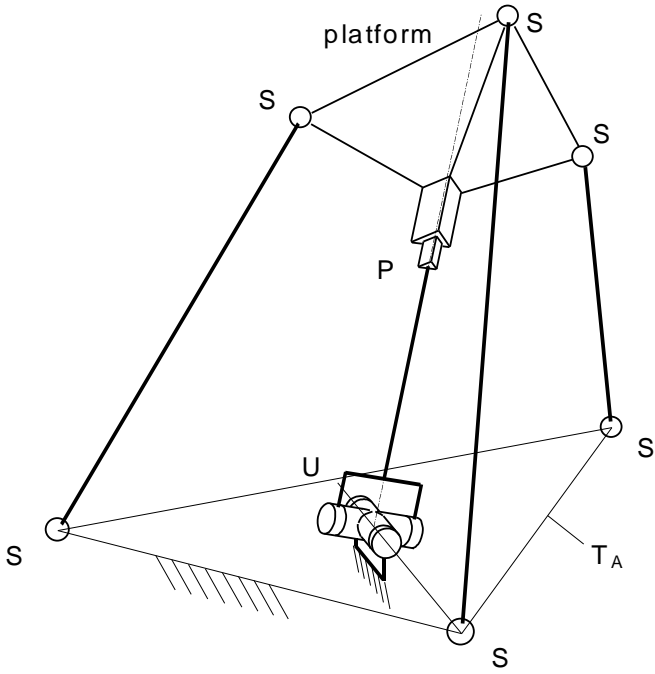

**Figure 3** – The UP-3(SS) structure associated with a Tricept manipulator (type-I).

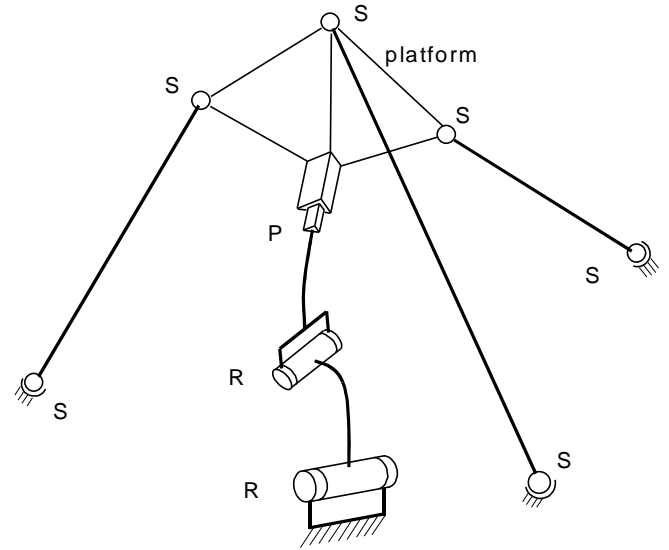

**Figure 4** – The general-geometry RRP-3(SS) structure.

Contrary to the convictions expressed in [3], the literature available at the time did not provide any means to solve exhaustively the forward kinematics of any Tricept manipulator. It were Joshi and Tsai [6] who first presented the forward kinematics solution for a Tricept, precisely the Tricept with the U-joint of the UP sub-chain having the axis of the fixed-axis revolute pair going through the center of a U-joint of a UPS sub-chain. For such a manipulator, 24 assembly configurations were recognized as possible in the complex domain. Moreover, in [6] the number of solutions for the forward kinematics of a general-geometry Tricept manipulator was predicted as high as 32 in the complex domain. Considering that the forward kinematics of a general-geometry Tricept can be solved by finding the assembly configurations of a special-geometry RRP-3(SS) structure, the procedure presented in this paper lowers such an esteem to 28.

The paper concludes with two numerical examples that show application of the presented results to as many case studies.

## COMPATIBILITY EQUATIONS

The position analysis of the general-geometry RRP-3(SS) structure will be carried out by solving a set of three compatibility equations in a corresponding number of unknowns. The present section is devoted to showing the procedure conducive to finding the above-mentioned set of equations. Actual solution of these equations is deferred to the next section.

In order to determine a set of compatibility equations, the three SS links of the RRP-3(SS) structure shown in Fig. 4 are thought of as temporarily removed so as to obtain a general-geometry RRP open-loop mechanism. The generic configuration of this RRP mechanism can be expressed in terms of three parameters by referring to the mechanism's reference (or home) configuration shown in Fig. 5 and specified in detail hereafter. The compatibility equations will be derived by writing the position of point Q, attached to link 3, as function of the abovementioned three parameters. For ease of subsequent reference, the four links of the RRP mechanism are sequentially labeled from 0 to 3 starting from the fixed link (base).

As a first step, a Cartesian reference frame $O_b x_b y_b z_b$ fixed to the base of the RRP mechanism is set up with its origin, $O_b$, at the point where the axis of the revolute pair between links 0 and 1 meets the common perpendicular to the axes of the two revolute pairs (see Fig. 5). Such a reference frame is chosen with axis $x_b$ superimposed on the axis of the revolute pair between links 0 and 1. Let $\mathbf{n_1}$ be a unit vector that has the same direction and orientation as axis $x_b$. The orientation of reference frame $O_b x_b y_b z_b$ about $\mathbf{n_1}$ with respect to link 0 can be chosen arbitrarily.

At the home configuration of the RRP mechanism, the position of link 1 is such as to make the shortest line segment between the axes of the two revolute pairs, $O_b Q$, overlap axis $z_b$ of reference frame $O_b x_b y_b z_b$ (see Fig. 5). Now let $\mathbf{n_2}$ be a unit vector along the axis of the revolute pair between links 1 and 2. The magnitude $\alpha$ of the hypothetical rotation about axis $z_b$ – positive if counterclockwise – that would make unit vector $\mathbf{n_1}$ align with unit vector $\mathbf{n_2}$ is a geometric parameter of the RRP mechanism. A further geometric parameter is the z-coordinate, $\zeta$, of point Q with respect to reference frame $O_b x_b y_b z_b$.

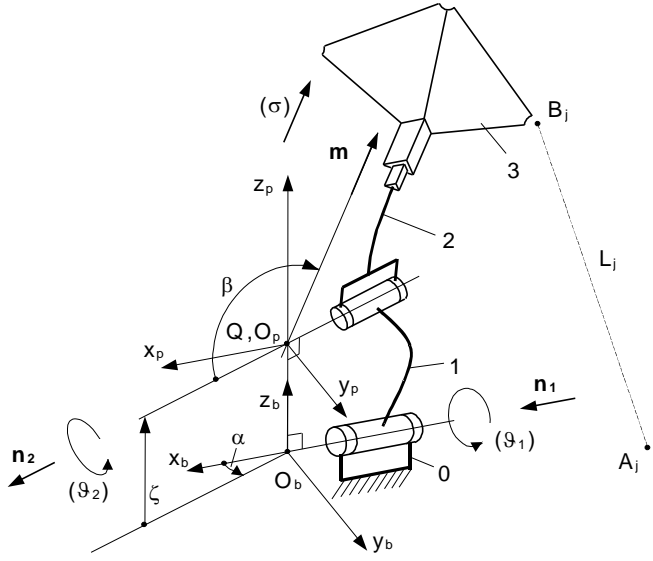

**Figure 5** – The RRP mechanism at the home configuration.

A unit vector, **m**, is now introduced parallel to the direction of the sliding motion of link 3 relative to link 2. At the home configuration of the RRP mechanism, the location of links 2 and 3 is so chosen as to make unit vector **m** become parallel to the plane defined by unit vector $\mathbf{n_2}$ and axis $z_b$ of reference frame $O_b x_b y_b z_b$. The angle β between unit vectors $\mathbf{n_2}$ and **m** is the last geometric parameter of the RRP mechanism (β is conventionally restricted to the range ]–π, π] and is considered as positive if **m** has a positive component along axis $z_b$). The definition of the home configuration of the RRP mechanism is concluded with the arbitrary selection of a reference position for link 3 relative to link 2.

With the RRP mechanism at the home configuration, a reference frame $O_p x_p y_p z_p$ fixed to link 3 is introduced parallel to $O_b x_b y_b z_b$ and with origin at point Q. Starting from the home configuration of the RRP mechanism, any location allowed to link 3 in space can be thought of as reached by the following elemental maneuvers, executed in any order (see also Fig. 5):

– a translation of magnitude σ of link 3 with respect to link 2 along the direction defined by unit vector **m** (considered as fixed to link 2 or 3);
– a rotation of magnitude $\vartheta_2$ of links 2 and 3 about the axis through point Q and parallel to the direction defined by unit vector $\mathbf{n_2}$ (considered as fixed to link 1 or 2);
– a rotation of magnitude $\vartheta_1$ of links 1, 2, and 3 about the x-axis of reference frame $O_b x_b y_b z_b$.

The location of link 3 relative to link 0 can be expressed in terms of both the 3×3 orthogonal matrix **R** for transformation of components of vectors from $O_p x_p y_p z_p$ to $O_b x_b y_b z_b$ and the coordinates of point Q with respect to reference frame $O_b x_b y_b z_b$. Matrix **R** is given by

$$\mathbf{R} = \mathbf{R_1} \mathbf{R_2} \tag{1}$$

where $\mathbf{R_i}$ (i=1, 2) is the 3×3 orthogonal matrix that represents a rotation $\vartheta_i$ about an axis parallel to unit vector $\mathbf{n_i} = (n_{ix}, n_{iy}, n_{iz})^T$. The detailed expression of $\mathbf{R_i}$ is (see, for example, [7])

$$\mathbf{R_i} = c_i \mathbf{I} + (1 - c_i) \mathbf{n_i} \mathbf{n_i}^T + s_i \begin{pmatrix} 0 & -n_{iz} & n_{iy} \\ n_{iz} & 0 & -n_{ix} \\ -n_{iy} & n_{ix} & 0 \end{pmatrix} \tag{2}$$

where $c_i = \cos \vartheta_i$, $s_i = \sin \vartheta_i$, and **I** is the 3×3 identity matrix. In using Eqs. (1) and (2), the following relations have to be taken into account

$$\mathbf{n_1} = \begin{pmatrix} 1 \\ 0 \\ 0 \end{pmatrix}; \quad \mathbf{n_2} = \begin{pmatrix} \cos \alpha \\ \sin \alpha \\ 0 \end{pmatrix} \tag{3}$$

The position of point Q with respect to reference frame $O_b x_b y_b z_b$ is given by

$$(Q - O_b) = \zeta \mathbf{R_1} \mathbf{k} + \sigma \mathbf{R_1} \mathbf{R_2} \mathbf{m} \tag{4}$$

where $\mathbf{k} = (0, 0, 1)^T$ is the unit vector along the z-axis of reference frame $O_b x_b y_b z_b$ and **m** – an already-introduced unit vector – has the following expression

$$\mathbf{m} = \begin{pmatrix} \cos \alpha \cos \beta \\ \sin \alpha \cos \beta \\ \sin \beta \end{pmatrix} \tag{5}$$

Equations (1) and (4) together express the rigid-body position of link 3 of the RRP mechanism in terms of parameters $\vartheta_1$, $\vartheta_2$, and σ.

Although a generic choice for parameters $\vartheta_1$, $\vartheta_2$, and σ certainly results into an admissible configuration of the RRP mechanism, the same choice does not necessarily correspond to an assembly configuration of the RRP-3(SS) structure.

To solve the position analysis of the RRP-3(SS) structure, the formerly-removed SS-links are now ideally added to the RRP mechanism, thus re-obtaining the original RRP-3(SS) structure. Insertion of the three SS-links into the RRP mechanism is feasible if and only if the following set of compatibility equations is satisfied

$$\left( B_j - A_j \right)^2 = L_j^2 \qquad (j = 1, 2, 3) \tag{6}$$

In Eqs. (6), $A_j$ and $B_j$ (j=1,2,3) are the centers of the spherical pairs that connect the generic SS-link of the RRP-3(SS) structure to base and platform respectively, whereas $L_j$ is the length of the considered SS-link (see fig. 5).

In order to implement Eqs. (6), the components with respect to $O_b x_b y_b z_b$ of vector $(B_j - A_j)$ are expressed in terms of parameters $\vartheta_1$, $\vartheta_2$, and $\sigma$ by taking advantage of Eqs. (1) and (4)

$$(B_j - A_j) = \zeta \mathbf{R}_1 \mathbf{k} + \mathbf{R}_1 \mathbf{R}_2 (\sigma \mathbf{m} + \mathbf{b}_j) - \mathbf{a}_j \qquad (7)$$
$$(j = 1, 2, 3)$$

where $\mathbf{a}_j$ and $\mathbf{b}_j$ are the vector of coordinates of points $A_j$ and $B_j$ with respect to reference frames $O_b x_b y_b z_b$ and $O_p x_p y_p z_p$ respectively. Algebraic manipulation of Eq. (6) leads to

$$\mathbf{a}_j^T \mathbf{R}_1 \mathbf{R}_2 (\sigma \mathbf{m} + \mathbf{b}_j) + \zeta \mathbf{a}_j^T \mathbf{R}_1 \mathbf{k} - \zeta \mathbf{k}^T \mathbf{R}_2 (\sigma \mathbf{m} + \mathbf{b}_j)$$
$$- \sigma \mathbf{m}^T \mathbf{b}_j - \frac{1}{2}(\sigma^2 + \mathbf{a}_j^2 + \mathbf{b}_j^2 + \zeta^2 - L_j^2) = 0 \qquad (8)$$
$$(j = 1, 2, 3)$$

Equations (8) is a set of three conditions in three unknowns, namely $\vartheta_1$, $\vartheta_2$, and $\sigma$ (matrices $\mathbf{R}_1$ and $\mathbf{R}_2$ depend on $\vartheta_1$ and $\vartheta_2$ respectively, see Eq. (2)). The generic solution of equation set (8) represents an assembly configuration of the RRP-3(SS) structure because it directly provides the configuration of the corresponding RRP mechanism that is compatible with the presence of the three SS-links.

A procedure to solve exhaustively Eqs. (8) is dealt with in the next section.

## ALGEBRAIC ELIMINATION AND BACK-SUBSTITUTION

The determination of all solutions of equation set (8) is carried out by first making algebraic the dependence of all equations on the unknowns. After algebraic elimination, a univariate algebraic equation will be determined whose roots will provide the sought-for assembly configurations of the RRP-3(SS) structure.

The left-hand sides of Eqs. (8) depend linearly on matrices $\mathbf{R}_i$ (i=1,2), which in turn depend at most linearly on either $s_i = \sin\vartheta_i$ or $c_i = \cos\vartheta_i$ (see Eq. (2)). In addition, almost any term on the left-hand side of Eqs. (8) depends at most linearly on unknown $\sigma$, the only exception being the first term within the last pair of parentheses on the left-hand side of Eqs. (8) (whose coefficient is constant). If the first of Eqs. (8) is subtracted from the remaining ones, a new equation set can be obtained as formed by the first of Eqs. (8) together with the two just-obtained equations. Such an equation set can be synthetically written as

$$\sum_{\substack{p,q,u,v=0,1 \\ p+q \leq 1 \\ u+v \leq 1}} e_{jpquv}(\sigma) s_1^p c_1^q s_2^u c_2^v = 0 \qquad (j = 1, 2, 3) \qquad (9)$$

where quantities $e_{jpquv}(\sigma)$ depend at most linearly on $\sigma$ for j=2,3. For the first of Eqs. (9) – j=1 – the dependence of $e_{jpquv}(\sigma)$ on $\sigma$ is at most linear as soon as at least one of indices p, q, u, and v is different from zero; at most quadratic if all of these indices vanish.

By replacing the sine and cosine of $\vartheta_i$ (i=1,2) with the following expressions in terms of $t_i = \tan(\vartheta_i/2)$

$$s_i = \frac{2t_i}{1+t_i^2}; \qquad c_i = \frac{1-t_i^2}{1+t_i^2} \qquad (i = 1, 2) \qquad (10)$$

Eqs. (9) can be re-written – after rationalization – in the ensuing form

$$\sum_{\substack{u,v=0,2 \\ w=0,d(j)}} f_{juvw} t_1^u t_2^v \sigma^w = 0 \qquad (j = 1, 2, 3) \qquad (11)$$

In Eqs. (11), the upper bound for index w, d(j), equals 2 for j=1, equals 1 for j=2,3. Coefficients $f_{juvw}$ (j=1,2,3; u,v,w=0,1,2) appearing in Eqs. (11) are constant quantities that depend on the geometry of the RRP-3(SS) structure only. Their expression is here omitted for the sake of brevity.

With the aim at eliminating unknowns $t_1$ and $t_2$ from Eqs. (11), these same equations can be re-written as

$$\sum_{p,q=0,2} f_{jpq}(\sigma) t_1^p t_2^q = 0 \qquad (j = 1, 2, 3) \qquad (12)$$

where quantities $f_{juv}(\sigma)$ (p,q=0,1,2) are second-order or first-order polynomials in unknown $\sigma$ depending on whether j=1 or j=2, 3 respectively.

The same process of algebraic elimination shown in [8] is now adopted. Starting from each of Eqs. (12), auxiliary equations are obtained through multiplication by factors $t_1^u t_2^v$ (u = 0,..,3; v = 0,1). As a consequence, the j-th equation of equation set (12) generates eight auxiliary equations and the whole set of auxiliary equation stemming from Eqs. (12) is

$$\sum_{p,q=0,2} f_{juv}(\sigma) t_1^{p+u} t_2^{q+v} = 0$$
$$(u = 0,..,3; v = 0,1; j = 1, 2, 3) \qquad (13)$$

The left-hand sides of these equations can in turn be regarded as linearly dependent on the twenty-four power products $t_1^u t_2^v$ (u = 0,..,5; v = 0,..,3) and arranged in matrix form as follows

$$\mathbf{M} \boldsymbol{\tau} = \mathbf{0} \qquad (14)$$

In this equation, $\mathbf{M}$ is a 24×24 matrix whose elements depend on unknown $\sigma$, whereas $\boldsymbol{\tau}$ is the following 24-component vector

$$\boldsymbol{\tau} = \begin{pmatrix} t_1^0 t_2^0 & t_1^1 t_2^0 & t_1^2 t_2^0 & t_1^3 t_2^0 & t_1^4 t_2^0 & t_1^5 t_2^0 & t_1^0 t_2^1 & t_1^1 t_2^1 & t_1^2 t_2^1 \\ & & & & \ldots & t_1^5 t_2^3 \end{pmatrix}^T \qquad (15)$$

Equation (14) is now regarded as a linear, homogeneous set of 24 equations in the 24 power products. Since it must be satisfied at an assembly configuration of the RRP-3(SS) structure and one of the power products is known *a priori* to be different from 0 ($t_1^0 t_2^0 = 1$), matrix **M** must be singular at an assembly configuration of the RRP-3(SS) structure, which leads to the ensuing condition

$$\det \mathbf{M} = 0 \qquad (16)$$

Equation (16) itself is the outcome of elimination of unknowns $t_1$ and $t_2$ from the set of compatibility equations.

Owing to the dependence of the elements of matrix **M** on unknown σ − at most quadratic for the elements on the first eight rows of **M**, at most linear for the elements on the remaining rows − the degree of the left-hand side of Eq. (16) is expected not to exceed 32. Actually, direct verification by algebraic manipulation software − employed to carry out exact-arithmetic computations for a number of RRP-3(SS) structures with randomly-chosen geometric parameters − has consistently shown that the left-hand side of Eq. (16) always reduces to a 28[th]-order polynomial in σ. Consequently Eq. (16) itself can be re-written as

$$\sum_{w=0}^{28} h_w \, \sigma^w = 0 \qquad (17)$$

where coefficients $h_w$ (w=0,..,28) depend on the geometry of the RRP-3(SS) only.

In the complex domain, Eq. (17) admits twenty-eight roots that can be determined by resorting to well-known numerical algorithms.

For the generic solution $\sigma_j$ ($1 \leq j \leq 28$) of Eq. (17), the corresponding values $t_{1j}$ and $t_{2j}$ of unknowns $t_1$ and $t_2$ can be found by back-substitution, which specifically consists in evaluating matrix **M** for the considered value $\sigma_j$ of σ, linearly solving Eq. (14) with the provision that the first component of vector **τ** is unitary, and selecting the second and seventh components of vector **τ** (see Eq. (15)). Once $t_{1j}$ and $t_{2j}$ have been determined, Eqs. (10) directly provide the sine and cosine of $\vartheta_{1j}$ and $\vartheta_{2j}$ which, together with $\sigma_j$, allow determination of the j-th assembly configuration of the RRP-3(SS) structure through Eqs. (1) and (4).

**APPLICATION TO THE FORWARD KINEMATICS OF TRICEPT MANIPULATORS**

As already stated in the introductory section of this paper, the procedure that has been proposed for carrying out the position analysis of the RRP-3(SS) structure can be specialized to solve the forward kinematics of the 3-dof in-parallel-actuated elemental manipulator of a Tricept manipulator. Due to the geometric specificities of such an elemental manipulator, the corresponding UP-3(SS) structure is associated with an RRP mechanism (see Fig. 5) characterized by ζ = 0 (which means $O_p \equiv O_b$), α = π/2, and β = π/2.

For a Tricept manipulator, the UP-3(SS) structure if further characterized by having:
− points $A_j$ (j=1,..,3) at the vertices of an equilateral triangle, $T_A$;

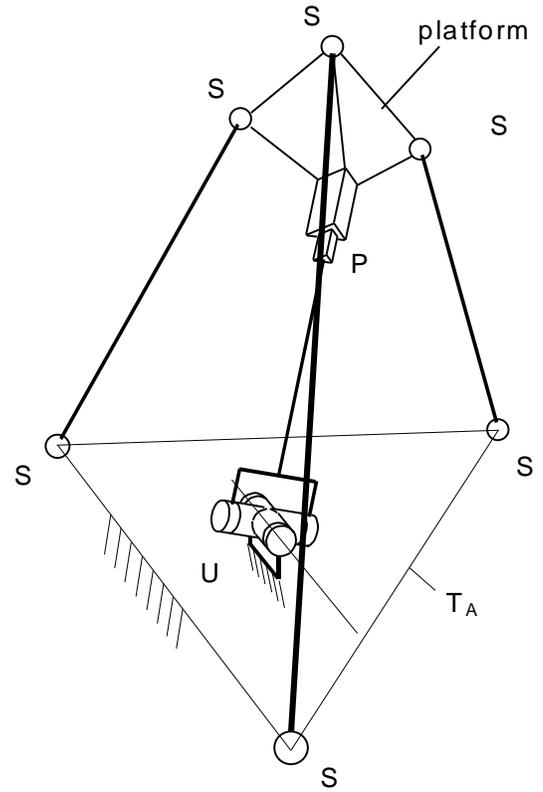

**Figure 6** – The UP-3(SS) structure associated with a Tricept manipulator (type-II).

− unit vector $\mathbf{n}_1$ parallel to the plane that goes through the vertices of triangle $T_A$;
− point $O_b$ at the center of triangle $T_A$;
− points $B_j$ (j=1,..,3) at the vertices of an equilateral triangle, $T_B$;
− unit vector **m** orthogonal to triangle $T_B$;
− points $B_j$ (j=1,..,3) equidistant from point $O_p$;
− the sides of triangle $T_B$ parallel to the sides of triangle $T_A$ at the home configuration of the underlying RRP mechanism.

It has been already mentioned in the introductory section that there exist two different types of in-parallel-actuated elemental manipulators for Tricept manipulators. The corresponding UP-3(SS) structures − shown in Figs. 3 and 6 and here dubbed as type-I and type-II structures respectively − differ in the placement with respect to triangle $T_A$ of the revolute pair axis parallel to unit vector $\mathbf{n}_1$ (see Figs. 5, 3, and 6): such an axis goes through a vertex of triangle $T_A$ for a type-I UP-3(SS) structure (as for the Tricept 805), whereas it is parallel to a side of triangle $T_A$ for a type-II UP-3(SS) structure (as for the Tricept 600).

The procedure presented in this paper for the solution of the position analysis of the RRP-3(SS) structure can be specialized to tackle the forward kinematics of Tricept manipulators whose in-parallel-actuated elemental manipulator involves a type-II UP-3(SS) structure. Twenty-eight assembly configurations are possible for this structure in the complex domain. Due to geometric specificities of the type-II structure, all odd-index coefficients of Eq. (17) vanish, which means that Eq. (17) itself can be regarded as a 14[th]-order algebraic equation in the unknown $\sigma^2$. This happens because a type-II UP-3(SS) structure can be assembled in either of two configurations that are symmetric with respect to the plane that contains the base points $A_i$ (i=1,2,3) and the U-joint center.

The forward kinematics of Tricept manipulators that involve a type-I UP-3(SS) cannot be solved by the procedure presented in this paper. The reason has to be sought in the algebraic elimination procedure here adopted, which does not profit from the simplified form taken in this case by the compatibility equations. The reader is therefore referred to the procedure explained in [6], where the 24 solutions that exist in the complex domain are determined by solving a 12$^{th}$-order algebraic equation in the square of one of the unknowns, followed by extraction of square root and linear back-substitution.

**NUMERICAL EXAMPLES**

This section shows application of the presented method to the position analysis of two structures: the first one is an RRP-3(SS) structure, whereas the other is a type-II UP-3(SS) structure.

With reference to Fig. 5, the coordinates of points $A_i$ (i=1,..,3) of an RRP-3(SS) structure with respect to the base reference frame $O_b x_b y_b z_b$ are: $A_1=(-1, 2, -1)^T$, $A_2=(-1, -1, 1)^T$, $A_3=(2, 0, 2)^T$ (throughout this section, all linear dimensions can be regarded as expressed in the same arbitrary length unit; the input data are to be considered as exactly defined by the reported digits; the numerical results are provided with a relatively high number of decimal digits – all meaningful – in order to enable the reader to verify their correctness with confidence). The coordinates of points $B_i$ (i=1,..,3) with respect to the platform reference frame $O_p x_p y_p z_p$ are: $B_1=(-1, 1, 0)^T$, $B_2=(0, -1, 1)^T$, $B_3=(1, -1, 1)^T$. The remaining geometric parameters of the considered RRP-3(SS) structure are: $\alpha=80°$, $\beta=115°$, $\zeta=1$, $L_1=3$, $L_2=4$, $L_3=5$. By following the procedure previously outlined, a polynomial equation of 28$^{th}$-order in the unknown $\sigma$ is found that has all coefficients different from zero. Its twenty-eight roots, $\sigma_i$ (i=1,..,28), are listed in Table 1; eight of them are real, whereas the remaining are complex. Therefore the considered RRP-3(SS) structure can actually be assembled in eight different ways. For any real value of $\sigma_i$ reported in Table 1, Table 2 lists the corresponding values of $\sigma$, $\vartheta_1$, $\vartheta_2$, and the coordinates of points $B_i$ (i=1,..,3) with respect to reference frame $O_b x_b y_b z_b$. It can be easily checked that the positions of points $B_i$ (i=1,..,3) are compatible – for each assembly configuration – with the lengths $L_i$ (i=1,..,3) of the three SS links.

As a second example, a type-II UP-3(SS) structure is now considered. By referring again to Fig. 5, the coordinates of points $A_i$ and $B_i$ (i=1,..,3) with respect to $O_b x_b y_b z_b$ and, respectively, $O_p x_p y_p z_p$ are $A_i=4\,\mathbf{u_i}$ and $B_i=3\,\mathbf{u_i}$, where $\mathbf{u_i}$ (i=1,..,3) is given by

$$\mathbf{u_i} = \left(\cos\left(i\,120°-150°\right),\quad \sin\left(i\,120°-150°\right),\quad 0\right)^T \quad (18)$$

**Table 1** – The roots $\sigma_i$ (i=1,..,28) of Eq. (17) (real and imaginary parts) for the considered RRP-3(SS) structure.

| Root # | Root value | |
|---|---|---|
| 1 | ( −5.0742351861635417, | 0. ) |
| 2 | ( −4.9208457694073359, | 0. ) |
| 3 | ( −3.2485304798567102, | 0. ) |
| 4 | ( −2.9472972942348737, | 0. ) |
| 5 | ( 0.4336937265758375, | 0. ) |
| 6 | ( 1.8716859056627936, | 0. ) |
| 7 | ( 2.8533551381339947, | 0. ) |
| 8 | ( 3.0202234858973762, | 0. ) |
| 9-10 | ( −2.6539388259158195, | ±0.3470682923497006 ) |
| 11-12 | ( 0.6288934148939096, | ±0.0920713380338177 ) |
| 13-14 | ( 5.3978372439452376, | ±1.6353960015160476 ) |
| 15-16 | ( −1.0796669034069113, | ±0.3665477831699458 ) |
| 17-18 | ( −1.1754717456325313, | ±0.5718777661241322 ) |
| 19-20 | ( −0.5607303198355512, | ±0.2771024466996316 ) |
| 21-22 | ( −1.7925839699411944, | ±0.9012404143420023 ) |
| 23-24 | ( −5.1649516067821035, | ±2.7634387909159706 ) |
| 25-26 | ( 2.2577581202525811, | ±1.5176837470894034 ) |
| 27-28 | ( 0.1539845408833452, | ±1.5772504431994877 ) |

Parameters $\alpha$, $\beta$, and $\zeta$ are now mandatorily set to $\alpha=90°$, $\beta=90°$, $\zeta=0$. The lengths of the three SS links are chosen as $L_1=6$, $L_2=7$, $L_3=7$. The 28$^{th}$-order polynomial equation in unknown $\sigma$ – which is now deprived of the odd-power terms – is satisfied by the fourteen values for $\sigma^2$ listed in Table 3, six of which are real and positive. Consequently, Eq. (17) has twelve real solutions, as many as are the square roots (with both plus and minus signs) of the aforementioned positive real values for $\sigma^2$. The corresponding twelve real assembly configurations of the considered type-II UP-3(SS) structure are listed in Table 4 in terms of the values of parameters $\sigma$, $\vartheta_1$, $\vartheta_2$ and the coordinates of points $B_i$ (i=1,..,3) with respect to reference frame $O_b x_b y_b z_b$. (In Table 4, pairs of assembly configurations that are symmetric with respect to the x-y plane of reference frame $O_b x_b y_b z_b$ are listed together; the selection of a single configuration is made by choosing the upper or lower signs wherever such an alternative exists.) Once more, it is easy to verify that the computed positions of points $B_i$ (i=1,..,3) are congruent with the lengths $L_i$ (i=1,..,3) of the three SS links.

**CONCLUSIONS**

A procedure to exhaustively solve the position analysis of a general-geometry RRP-3(SS) multi-loop spatial structure has been presented. The twenty-eight solutions that exist in the complex domain can be determined by finding the roots of a univariate algebraic equation of 28$^{th}$ order, followed by linear back-substitutions.

If suitably specialized, the proposed procedure can find application to some types of Tricept manipulators, specifically those whose forward kinematics falls outside the reach of an already-known solving method.

Numerical examples have shown application of the presented results to two case studies.

**Table 2** – Values of σ, $\vartheta_1$ and $\vartheta_2$, together with (x,y,z)-cooordinates of points $B_i$ (i=1,2,3) relative to reference frame $O_b x_b y_b z_b$, for the real assembly configurations of the considered RRP-3(SS) structure.

___________________________________________________________________________________

| # | | | | |
|---|---|---|---|---|
| 1 | (σ, $\vartheta_1$, $\vartheta_2$) | ( −5.0742351861635417, | 35.9079893748161347°, | 28.9649324307956022° ) |
| | $B_1$ | ( −2.6781700217812648, | 4.2576192315137761, | 0.0425453388192841 ) |
| | $B_2$ | ( −1.3653461635426380, | 2.4582409822157228, | −0.9766364080701107 ) |
| | $B_3$ | ( −0.4866581591368590, | 2.7552713092491167, | −1.3503747868002389 ) |
| 2 | (σ, $\vartheta_1$, $\vartheta_2$) | ( −4.9208457694073359, | −16.7397063715162090°, | 9.9331724537507540° ) |
| | $B_1$ | ( −1.3793980152669597, | 2.1242250978426177, | −3.9733188983150603 ) |
| | $B_2$ | ( −0.2291846275909712, | 0.4011369623340533, | −2.6664231717628057 ) |
| | $B_3$ | ( 0.7562770072030819, | 0.3546626347909773, | −2.8298413568844473 ) |
| 3 | (σ, $\vartheta_1$, $\vartheta_2$) | ( −3.2485304798567102, | −79.0280445391782827°, | 163.9997204883860072° ) |
| | $B_1$ | ( 0.6767083869286662, | 4.4729707002083387, | −1.2703051999391603 ) |
| | $B_2$ | ( −0.6247488516958736, | 2.8653898224331031, | 0.0419039211399423 ) |
| | $B_3$ | ( −1.5268699670835949, | 2.6627327388291424, | −0.3390259931970009 ) |
| 4 | (σ, $\vartheta_1$, $\vartheta_2$) | ( −2.9472972942348737, | −96.1096693511796809°, | 174.4317612902740866° ) |
| | $B_1$ | ( 1.2373743598687456, | 3.5504833066462911, | −2.2610144684392421 ) |
| | $B_2$ | ( −0.2846112277092414, | 2.5978678155458440, | −0.5948561334301278 ) |
| | $B_3$ | ( −1.2197274745538063, | 2.4665375488648706, | −0.9239609282004854 ) |
| 5 | (σ, $\vartheta_1$, $\vartheta_2$) | ( 0.4336937265758375, | 170.8277016071986500°, | −12.7989139878393903° ) |
| | $B_1$ | ( −1.0892329362024957, | −0.9986059923310343, | −0.9800103563539957 ) |
| | $B_2$ | ( −0.3399926313342539, | 0.7296870316619115, | −2.5457824306924873 ) |
| | $B_3$ | ( 0.6359101341710010, | 0.6907160998793561, | −2.7604800290842598 ) |
| 6 | (σ, $\vartheta_1$, $\vartheta_2$) | ( 1.8716859056627936, | 80.9195928499276312°, | 169.0366603163963822° ) |
| | $B_1$ | ( 1.4412409995520388, | 0.4025929011426914, | −0.3009771515705376 ) |
| | $B_2$ | ( 0.0287424686051592, | 1.3736985290103308, | −2.0507776429399693 ) |
| | $B_3$ | ( −0.8932495553923580, | 1.6121281380216517, | −1.7456841667205158 ) |
| 7 | (σ, $\vartheta_1$, $\vartheta_2$) | ( 2.8533551381339947, | −42.5300309414956836°, | −45.9066707230024256° ) |
| | $B_1$ | ( −2.6914641610939969, | 1.3842880246672619, | 1.3999682821167531 ) |
| | $B_2$ | ( −2.7977917841657982, | 1.1712865124310922, | 3.8378617975019116 ) |
| | $B_3$ | ( −2.0927907285236484, | 1.6877341071135708, | 4.3239228980748094 ) |
| 8 | (σ, $\vartheta_1$, $\vartheta_2$) | ( 3.0202234858973762, | 155.8002697774543024°, | −167.5798330690447930° ) |
| | $B_1$ | ( 0.4535864782038204, | 1.2918626159811463, | 1.5269799753337874 ) |
| | $B_2$ | ( −1.3512558881446214, | 2.9478794426552221, | 1.5393210614318953 ) |
| | $B_3$ | ( −2.2682508421575309, | 2.5527403119887298, | 1.4846829636338322 ) |

**REFERENCES**

[1] Weck, M., and Staimer, M., 2002, "Parallel Kinematic Machine Tools – Current State and Future Potentials," *Annals of the CIRP,* **51**(2), Feb., pp. 671-683.

[2] Neumann, K.E., 1988, "Robot", U.S. Patent No. 4,732,525, Mar. 22.

[3] Siciliano, B., 1999, "The Tricept robot: Inverse kinematics, manipulability analysis and closed-form direct kinematics algorithm," *Robotica,* **17**, pp. 437-445.

[4] Merlet, J.-P., 2000, *Parallel robots,* Kluwer Academic Publisher, Dordrecht, NL.

[5] Hunt, K.H., 1978, *Kinematic Geometry of Mechanisms,* Clarendon Press, Oxford, UK.

[6] Joshi, S.A., and Tsai, L.W., 2003, "The Kinematics of a Class of 3-DOF 4-Legged Parallel Manipulators," *ASME Journal of Mechanical Design,* **125**(1), pp. 52-60.

[7] Craig, J.J., 1989, *Introduction to robotics: mechanics and control,* Addison-Wesley, Reading, MA, USA.

[8] Lin, W., Crane, C.D., and Duffy, J., 1994, "Closed-Form Forward Displacement Analysis of the 4-5 In-Parallel Platforms," *ASME Journal of Mechanical Design,* **116**(1), pp. 47-53.

**Table 3** – The values of $\sigma^2$ (real and imaginary parts) that satisfy Eq. (17) for the considered type-II UP-3(SS) structure.

| Value # | Value | |
|---|---|---|
| 1 | ( −5.8696327988584050, | 0. ) |
| 2 | ( −4.0239570540158663, | 0. ) |
| 3 | ( −3.3666899614601473, | 0. ) |
| 4 | ( −3.0563577788337002, | 0. ) |
| 5 | ( 0.4733932871332810, | 0. ) |
| 6 | ( 2.0748284313315206, | 0. ) |
| 7 | ( 2.1224196727429542, | 0. ) |
| 8 | ( 15.5595408347198758, | 0. ) |
| 9 | ( 21.0556791945148852, | 0. ) |
| 10 | ( 43.4967317928178336, | 0. ) |
| 11-12 | ( −2.9629094839493732, | ±14.2663109067628893 ) |
| 13-14 | (−23.3534016594300759, | ±29.6740259421679502 ) |

**Table 4** – Values of $\sigma$, $\vartheta_1$ and $\vartheta_2$, together with (x,y,z)-coordinates of points $B_i$ (i=1,2,3) relative to reference frame $O_b x_b y_b z_b$, for the real assembly configurations of the considered type-II UP-3(SS) structure.

| | | | | |
|---|---|---|---|---|
| 1-2 | ($\sigma$, $\vartheta_1$, $\vartheta_2$) | ( ± 0.6880358182051869, | ∓ 156.7136782148684357°, | ± 132.9139078387645247° ) |
| | $B_1$ | ( −1.2651245735830280, | 0.4403916388669646, | ± 2.7710843193352510 ) |
| | $B_2$ | ( 0.5039020492352966, | −2.9408258391083399, | ∓ 0.7556582534072386 ) |
| | $B_3$ | ( 2.2729286720536212, | 1.9448237362397493, | ∓ 0.7244647499004386 ) |
| 3-4 | ($\sigma$, $\vartheta_1$, $\vartheta_2$) | ( ± 1.4404264755035297, | ± 166.0952410961427079°, | ± 119.6888747109510109° ) |
| | $B_1$ | ( −0.0354627690969302, | 2.1698695743141532, | ± 2.5229421027453670 ) |
| | $B_2$ | ( 1.2513383830026394, | −2.7406464460835599, | ± 1.4134488813273061 ) |
| | $B_3$ | ( 2.5381395351022091, | 1.0851062606708437, | ∓ 1.8588223515805918 ) |
| 5-6 | ($\sigma$, $\vartheta_1$, $\vartheta_2$) | ( ± 1.4568526599292580, | ∓ 121.5113162764218017°, | ± 159.9432882232469948° ) |
| | $B_1$ | ( −1.9408845599906445, | −1.1423155877154994, | ± 2.4597767979219313 ) |
| | $B_2$ | ( 0.4996277511034456, | −2.7346975409071307, | ∓ 1.8423412123780870 ) |
| | $B_3$ | ( 2.9401400621975357, | 0.3769231127194106, | ± 1.5283733361001609 ) |
| 7-8 | ($\sigma$, $\vartheta_1$, $\vartheta_2$) | ( ± 3.9445583827242151, | ∓ 50.1598159353873538°, | ± 169.3917522904197658° ) |
| | $B_1$ | ( −1.8275071621977196, | −4.3052204648024363, | ∓ 1.6385466597818347 ) |
| | $B_2$ | ( 0.7261649628862395, | −1.0550573956600155, | ∓ 4.7873875102460014 ) |
| | $B_3$ | ( 3.2798370879701985, | −3.5707296855931061, | ∓ 1.0257187838385341 ) |
| 9-10 | ($\sigma$, $\vartheta_1$, $\vartheta_2$) | ( ± 4.5886467715999763, | ∓ 150.30166338368242480°, | ± 10.1346512335003609° ) |
| | $B_1$ | ( 3.3649661993974588, | 3.31437262447962296, | ∓ 2.7834539294781876 ) |
| | $B_2$ | ( 0.8074279219136203, | −0.36804010068563934, | ∓ 5.4100171746230568 ) |
| | $B_3$ | ( −1.7501103555702182, | 3.76736025607133018, | ∓ 3.5776793650095212 ) |
| 11-12 | ($\sigma$, $\vartheta_1$, $\vartheta_2$) | ( ± 6.5952052123355368, | ± 4.85386761100620255°, | ± 8.3399034085793430° ) |
| | $B_1$ | ( 3.5272052199421251, | −2.01488429854254115, | ± 5.9996496891666592 ) |
| | $B_2$ | ( 0.9566036312166227, | 2.43709147410222920, | ± 6.7559030804489217 ) |
| | $B_3$ | ( −1.6139979575088798, | −2.07865648248668464, | ± 6.7506243869579265 ) |